\documentclass{article}
\PassOptionsToPackage{numbers, compress}{natbib}
\usepackage[final]{neurips_2023}
\usepackage[utf8]{inputenc} 
\usepackage[T1]{fontenc}    
\usepackage{hyperref}       
\usepackage{url}            
\usepackage{booktabs}       
\usepackage{amsfonts}       
\usepackage{nicefrac}       
\usepackage{microtype}      
\usepackage{xcolor}         
\usepackage{caption}
\usepackage{subcaption}
\usepackage{wrapfig,lipsum,booktabs}
\usepackage{graphicx}
\usepackage{adjustbox}
\usepackage{multirow}
\usepackage{makecell}
\usepackage{amssymb}
\usepackage{pifont}
\usepackage[capitalize]{cleveref}
\usepackage{xspace}
\usepackage{comment}
\usepackage{wrapfig}
\usepackage{enumitem}
\usepackage[olditem,oldenum]{paralist}
\setdefaultleftmargin{.2cm}{.2cm}{}{}{}{}
\hypersetup{
    colorlinks=true,
    linkcolor=black,
    filecolor=black,
    urlcolor=magenta,
}
\crefname{section}{Sec.}{Secs.}
\Crefname{section}{Section}{Sections}
\Crefname{table}{Table}{Tables}
\crefname{table}{Tab.}{Tabs.}
\newcommand{\modelNameWithEmoji}{LENS\includegraphics[width=1em]{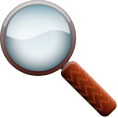}\xspace}
\newcommand{\modelName}{LENS\xspace}
\newcommand{\colorgreen}{\color[HTML]{3A9D1C}}
\newcommand{\colorred}{\color[HTML]{EE4B2B}}
\newcommand{\visionLanguage}{V\&L\xspace}
\definecolor{prompttext}{HTML}{666666}
\title{\Large~Towards Language Models That Can See: \\ Computer Vision Through the \modelNameWithEmoji of Natural Language }
\author{%
William Berrios$^{\dag}$ \quad Gautam Mittal$^{\dag\S}$  \quad Tristan Thrush$^{\dag\S}$ \\
\quad \textbf{Douwe Kiela$^{\dag\S}$} \quad \textbf{Amanpreet Singh$^{\dag}$ }\\
$^{\dag}$Contextual AI; $^{\S}$Stanford University\\
}

\begin{document}
\maketitle
\begin{abstract}
\label{sec:abstract}
We propose \modelNameWithEmoji, a modular approach for tackling computer vision problems by leveraging the power of large language models (LLMs). Our system uses a language model to reason over outputs from a set of independent and highly descriptive vision modules that provide exhaustive information about an image. We evaluate the approach on pure computer vision settings such as zero- and few-shot object recognition, as well as on vision and language  problems. \modelName can be applied to any off-the-shelf LLM and we find that the LLMs with \modelName perform highly competitively with much bigger and much more sophisticated systems, without any multimodal training whatsoever. We open-source our code at \url{https://github.com/ContextualAI/lens} and provide an interactive demo\footnote{\label{demo}\footnotesize \url{https://lens.contextual.ai/}}.
\end{abstract}





\begin{figure*}[!h]
  \centering
  \includegraphics[width=1\textwidth, angle=0]{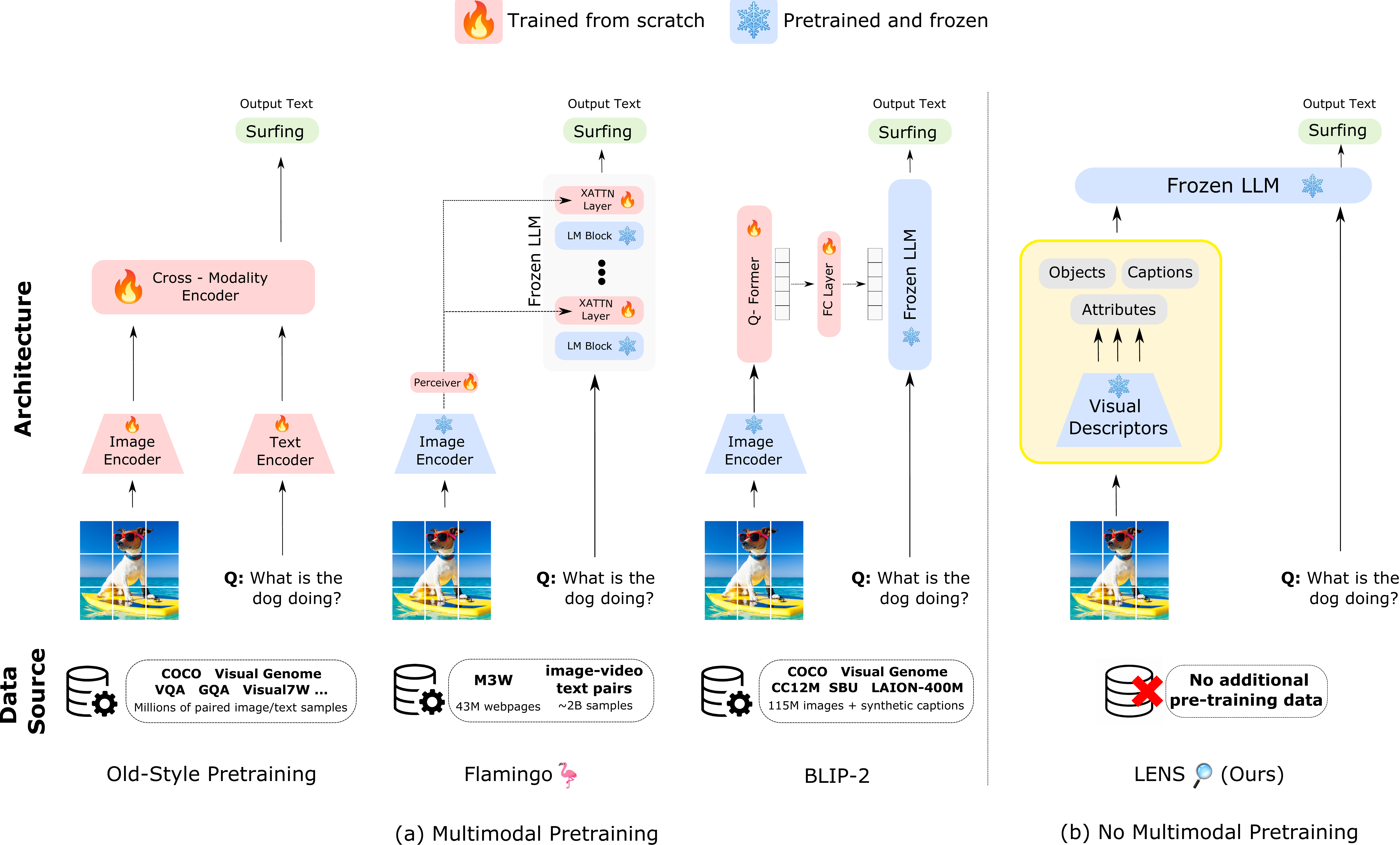}
  \caption{\textbf{Comparison of approaches for aligning visual and language modalities:} (a) Multimodal pretraining using a paired or web dataset, and (b) \modelNameWithEmoji, a pretraining-free method that can be applied to any off-the-shelf LLM without the need for additional multimodal datasets. Unlike \modelName, prior methods are computationally intensive and require joint alignment pretraining on large multimodal datasets to perform visual tasks.}
  \label{fig:teaser}
\end{figure*}
\section{Introduction}
\label{sec:intro}

In recent years, Large Language Models (LLMs) have revolutionized natural language understanding, showcasing remarkable capabilities in semantic understanding, question answering and text generation \cite{openai2019language,devlin2019bert,brown2020language}, especially in zero-shot and few-shot settings. Several approaches have been proposed for employing LLMs on vision-related tasks as shown in Fig.~\ref{fig:teaser}(a). One technique involves training a vision encoder to represent each image as a sequence of continuous embeddings, enabling comprehension by the LLM \cite{tsimpoukelli2021multimodal}. Another employs a frozen vision encoder that has been trained contrastively while introducing new layers into the frozen LLM that are subsequently trained from scratch \cite{radford2021learning, sung2022vladapter,alayrac2022flamingo}. Furthermore, an alternative approach suggests using both a frozen vision encoder (pre-trained contrastively) and a frozen LLM aligning them by training a lightweight transformer \cite{li2023blip2, koh2023grounding, zhu2023minigpt}.

While we have seen progress in the above research directions, the computational expense associated with the additional pretraining stage(s) still remains a challenge. Besides, large corpora of datasets containing images/videos and text are needed for aligning visual and language modalities on top of an existing LLM. An example of this is Flamingo \citep{alayrac2022flamingo} which introduces new cross-attention layers into an LLM to incorporate visual features, which are then pre-trained from scratch. Despite using a pretrained image encoder \cite{brock2021high} and a pretrained frozen LLM \cite{hoffmann2022training}, the multimodal pre-training stage still demands a staggering 2 billion image-text pairs along with 43 million webpages \cite{zhu2023multimodal,laurencon2023obe}, an undertaking that can last for approximately 15 days. Instead, as shown in Fig.~\ref{fig:teaser}(b), we can extract information from visual inputs and generate detailed textual representations (e.g. tags, attributes, actions, relationships, among others) using a diverse set of ``vision modules'' and then feed this information directly to the LLM avoiding the additional multimodal pretraining.

We introduce \modelNameWithEmoji (\textbf{L}arge Language Models \textbf{EN}nhanced to \textbf{S}ee) a modular approach that leverages a LLM as the ``reasoning module'' and operates over independent ``vision modules''. In the \modelName approach, we first extract rich textual information using pretrained vision modules such as contrastive models \citep{radford2021learning,singh2022flava,fang2022eva,brock2021high} and image-captioning models\citep{li2022blip,li2023blip2}. Subsequently, the text is fed into the LLM allowing it to perform object recognition and vision and language (\visionLanguage) tasks. \modelName eliminates the need for extra multimodal pretraining stages or data, bridging the gap between the modalities at zero cost. By integrating \modelName, we get a model which works across domains out of the box without any additional cross-domain pretraining \cite{jin2022good,hu2021unit,alayrac2022flamingo,li2023blip2}. Furthermore, this integration enables us to leverage the latest advancements in both computer vision and natural language processing out of the box, maximizing the benefits derived from these fields.

In summary, our contributions are as follows:
\begin{compactitem}

\item We propose \modelName, a modular approach that addresses computer vision tasks by harnessing the few-shot, in-context learning abilities of language models through natural language descriptions of visual inputs.
\item \modelName enables any off-the-shelf LLM to have visual capabilities without requiring auxiliary training or data. We utilize frozen LLMs to handle object recognition and visual reasoning tasks without the need for additional vision-and-language alignment or multimodal data.
\item Experimental results demonstrate that our approach achieves zero-shot performance that is competitive with or superior to end-to-end jointly pre-trained models like Kosmos and Flamingo.
\end{compactitem}

\section{Related Work}
\label{sec:related}
\subsection{Large Language Models capabilities}
LLMs have demonstrated remarkable abilities for natural language understanding and reasoning. GPT-3 \citep{brown2020language} is a notable example of such models, which can accurately solve complex tasks including translation, natural language inference, and common sense reasoning in a zero-shot or few-shot setting. Recently, more powerful versions such as GPT-3.5 and GPT-4 \citep{openai2023gpt4} were designed to understand, interact and generate human-like responses \cite{openai2021chatgpt}. These models are also known for their ability to perform a wide variety of tasks by showing a few examples in the prompt \cite{brown2020language}. Recent efforts have also been made to develop open-source LLMs that can compete with GPT-3, such as BLOOM \citep{bloom}, OPT \citep{zhang2022opt}, LLaMA \citep{touvron2023llama}, FlanT5 \citep{chung2022scaling} among others. However, all these models cannot directly solve tasks that require reasoning from a visual input stimulus. Our work leverages these LLMs as frozen language models and provides them with textual information obtained from the ``vision modules'' allowing them to perform object recognition and V\&L tasks.

\subsection{Contrastive Models for Solving Vision and Language tasks}
Foundation models such as \cite{radford2021learning,singh2022flava,jia2021scaling,fang2022eva,zeng2022multigrained} have demonstrated the ability to specify any visual concept based on an external vocabulary without the restriction of classes or labels presented in supervised models. However, previous work \citep{shen2021clip},~\cite{kim2021vilt} has shown that these contrastive models are unable to directly solve  tasks in zero or few shot settings. To address this, \cite{song2022clip} proposed a method using CLIP in VQA tasks by converting questions to a mask template that CLIP can answer, but their approach required fine-tuning for extending the model's capabilities to other tasks such as visual entailment \citep{xie2019visual}. In our work, we propose to leverage the capabilities of contrastive models and combine them with a crowdsourced open-source vocabulary to assign tags and attributes present in the image, which combined with frozen LLM can solve diverse \visionLanguage tasks.
\subsection{Large Language Models for Vision Applications}
\subsubsection{Image Captioning}
The field of image captioning has seen a significant surge in recent years, with the objective of generating natural language descriptions for images. To this end, various deep-learning models have been proposed. Notably, the recent models include BLIP\cite{li2022blip} and BLIP-2\cite{li2023blip2}, which achieve great performance on NoCaps\cite{Agrawal_2019} and COCO\cite{cocodataset}. Concurrently, ChatGPT has been leveraged to generate richer visual descriptions, along with BLIP-2 \cite{openai_gpt}. In another work, Socratic Models \cite{zeng2022socratic} and  Visual Clues \cite{xie2022visual} also use textual data to bridge the domain gap between vision-language models and language models. In particular, Visual Clues constructs a semantic representation of an image using structured textual prompts that include image tags, object attributes/locations, and captions. This approach leverages the GPT-3 large language model to generate image captions. Our work is inspired by Visual Clues, but instead of generating captions, we aim to utilize the raw compelling vision information with a frozen LLM in order to solve vision tasks.

\subsubsection{Vision and Language tasks}
LLMs can be leveraged in multiple ways in order to perform \visionLanguage task, these are mainly divided in two sections.

\textbf{Multimodal pretraining.} These approaches align vision and language modalities in different ways. For example, \citet{tsimpoukelli2021multimodal}, opts to finetune only the visual encoder and generate embeddings that are fed into a frozen LLM. Others, such as Flamingo \cite{alayrac2022flamingo}, train additional cross-attention layers for alignment. Works like BLIP2\cite{li2022blip} and Mini-GPT4\cite{zhu2023minigpt} reduce the size of extra layers and pretrained lightweight modules while freezing the vision encoder. However, in all cases, joint alignment of vision and language requires significant computing resources and training data, making it challenging to leverage state-of-the-art LLMs. Additionally, these approaches can hinder the reasoning abilities for which LLMs are renowned.

\textbf{Language-driven Modular Alignment}: These approaches couple LLMs with different modules in order to align the visual and language modalities. Concurrent work \citet{guo2023images} uses off-the-shelf LLMs for solving pure Visual Question Answering tasks like VQA 2.0 \citep{goyal2017making} and OK-VQA \citep{marino2019ok}. In contrast, \modelName extends the capacity of LLM to also solve object recognition tasks and also it does not involve any question-guided information extraction. Another work, PromptCap \citep{hu2023promptcap}, trains a question-aware captioning model using synthesized examples  with GPT-3 for solving VQA tasks. In contrast, \modelName  leverages ``vision modules'' without requiring any additional pre-training stage. Likewise, ViperGPT \citep{surís2023vipergpt} also leverages black box LLMs such as Instruct GPT and Codex to achieve great results on different VQA benchmarks but heavily relies on BLIP2 which needs extra training rounds of multimodal pre-training. Additionally, all the above methods rely on a ``top-down'' approach in which attention mechanisms are driven by nonvisual or task-specific contexts. However, our proposed approach differs from these methods as we utilize a ``bottom-up'' \citep{anderson2018bottomup} approach. Our method does not involve any question-guided information extraction, which is a more challenging task. Despite this, \emph{\modelName} achieves notable results that are comparable to these question-aware models.

\begin{figure*}[!h]
  \centering
  \includegraphics[width=1\textwidth, angle=0]{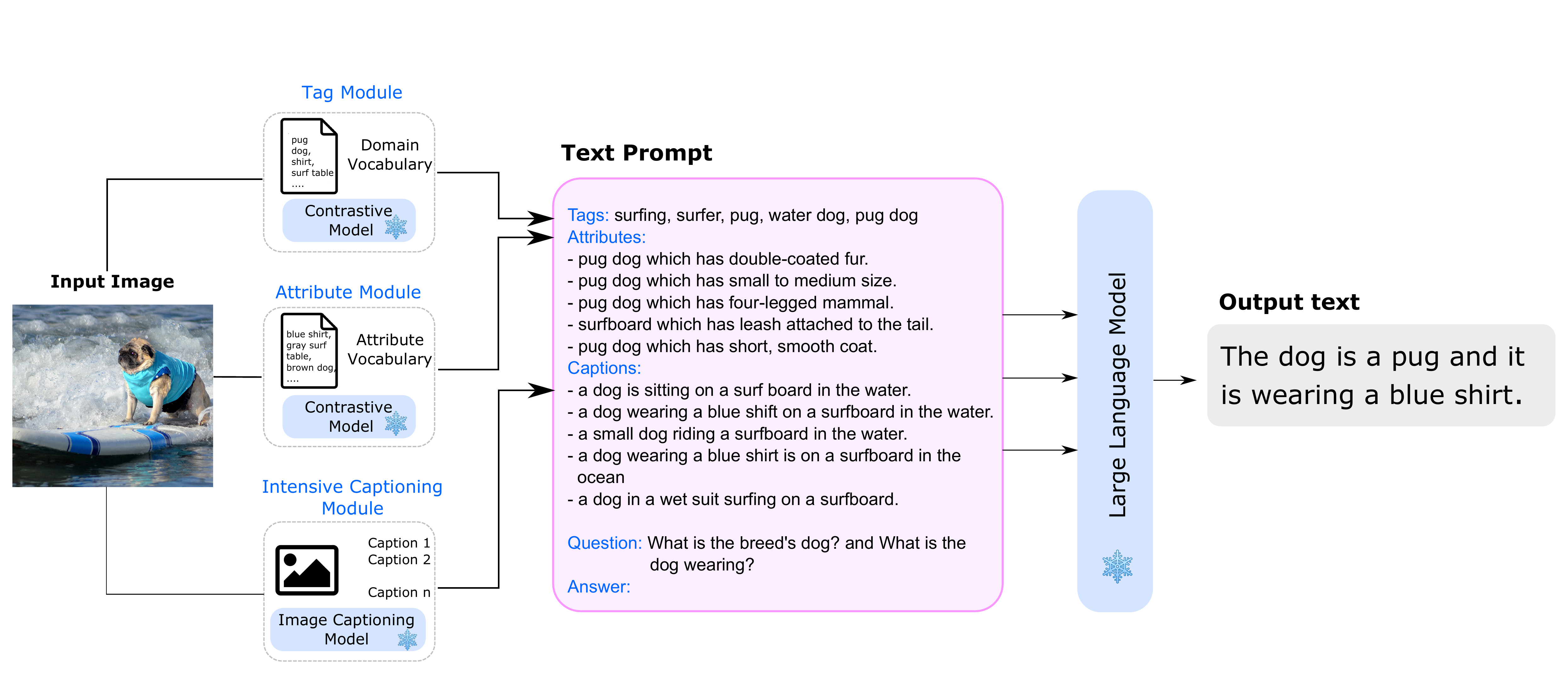}
  \caption{\textbf{The \modelName framework.} \modelName executes computer vision and visual reasoning tasks through a frozen LLM and a set of ``vision modules''. \modelName leverages these vision modules to retrieve a textual description for an image which is used by the ``reasoning module'' (LLM) to generate a response for a given query.}
  \label{fig:approach}
\end{figure*}
\section{Method}
\label{sec:method}
We present a novel framework called \modelNameWithEmoji (Fig.~\ref{fig:approach}), which aims to enhance the capabilities of frozen LLMs by enabling them to handle vision as well as vision-and-language tasks on top of their existing natural language understanding capabilities. In contrast to existing approaches, \modelName provides a unified framework that facilitates the operation of a LLM's ``reasoning module" on textual data extracted from a set of independent and highly descriptive ``vision modules''. More importantly, it eliminates the computational overhead of aligning the visual domain with the text domain through additional joint pretraining on multimodal data, a requirement in prior work for solving \visionLanguage tasks. \cite{alayrac2022flamingo, li2023blip2, zhu2023minigpt, gong2023multimodal, koh2023grounding}.

To summarize, given an image \emph{I}, we leverage the vision modules to extract all conceivable textual information \emph{T} that can describe the image, encompassing objects, attributes and captions, without limiting it to specific task instructions. Subsequently, a frozen LLM can process the generic prompts \emph{T} concatenated with task-specific prompts, and perform object recognition or visual reasoning tasks. In this section, we introduce the essentials of the ``vision modules", outline the main components of \modelName, and then discuss the prompt design.

\subsection{Visual Vocabularies}
\label{sec:visualvocabs}
For \modelName, visual vocabularies act as a bridge to convert an image into textual information which can then be handled by an existing LLM. We develop vocabularies for common objects and attributes.\\

\textbf{Tags}:
To create a diverse and comprehensive tag vocabulary for a contrastive model's image tagging, we collect tags from various sources. These include multiple image classification datasets such as ~\citep{imagenet,li_andreeto_ranzato_perona_2022,cimpoi14describing,parkhi2012pets,Nilsback08,bossard14,Xiao:2010,KrauseStarkDengFei-Fei_3DRR2013}, object detection and semantic segmentation datasets \citep{gupta2019lvis,cocodataset,OpenImages} along with the visual genome dataset \citep{krishnavisualgenome}.

\textbf{Attributes}:
Following the  methodology presented in~\citet{menon2022visual}, we employ a large language model, GPT-3, to generate descriptions of the visual characteristics that differentiate each object category within our object vocabulary.

\subsection{\modelNameWithEmoji Components}
\modelName consists of 3 distinct vision modules and 1 reasoning module, each serving a specific purpose based on the task at hand. These components are as follows:

\textbf{Tag Module.} Given an image, this module identifies and assigns tags to the image. To accomplish this, we employ a vision encoder (CLIP) that selects the most suitable tags for each image. In our work, we adopt a common prompt: \textcolor{prompttext}{\texttt{"A photo of \{classname\}"}} for object tagging in order to make our framework flexible across domains without the need for manual/ensemble prompt tuning \citep{radford2021learning}. We use the object vocabulary built in Section~\ref{sec:visualvocabs} as our class options.

\textbf{Attributes Module.} 
We utilize this module to identify and assign relevant attributes to the objects present in the image. For this purpose, we employ a contrastively pretrained vision encoder called CLIP, while incorporating the task-specific prompts outlined in \cite{menon2022visual}. The vision encoder classifies the objects based on the attributes vocabulary generated in Section~\ref{sec:visualvocabs}.


\textbf{Intensive Captioner.} We utilize an image captioning model called BLIP and apply stochastic top-k sampling \cite{fan-etal-2018-hierarchical} to generate N captions per image. This approach allows us to capture and encompass diverse aspects of the visual content within an image. These diverse captions are then directly passed to the "reasoning module" without any modifications.

\textbf{Reasoning Module.} We adopt a frozen LLM as our reasoning module, which is capable of generating answers based on the textual descriptions fed by the vision modules, along with the task-specific instructions. \modelName seamlessly integrates with any black box LLM, streamlining the process of adding vision capabilities to them and expediting the overall speed.

\subsection{Prompt Design}
With the textual information obtained from the vision modules, we construct complete prompts for the LLM by combining them. We formatted the tags module as \textcolor{prompttext}{\texttt{Tags: \{Top-k tags\}}},  the attributes modules as  \textcolor{prompttext}{\texttt{Attributes: \{Top-K attributes\}}}, the intensive captioning module as \textcolor{prompttext}{\texttt{Captions: \texttt{\{Top-N Captions\}}}}. In particular, for the hateful-memes task, we incorporate an OCR prompt as \textcolor{prompttext}{\texttt{OCR: this is an image with written "\{meme text\}" on it}}. Finally, we append the specific question prompt:  \textcolor{prompttext}{\texttt{Question: \{task-specific prompt\} \textbackslash n Short Answer:} } at the end. You can see this prompt in action in our demo\textsuperscript{\ref{demo}}.

\begin{table*}[t!]
\centering
\begin{tabular}{l@{\ }|cccc|cc}
\toprule
&  \multicolumn{4}{c}{LENS} & \multicolumn{2}{c}{CLIP}  \\
Datasets &  \small{L\textsubscript{14}- FlanT5\textsubscript{XL}} & \small{L\textsubscript{14}- FlanT5\textsubscript{XXL}} & \small{H\textsubscript{14}- FlanT5\textsubscript{XL}}& \small{H\textsubscript{14}- FlanT5\textsubscript{XXL}} & L\textsubscript{14} & H\textsubscript{14} \\
\midrule 
Pets \cite{parkhi2012pets}       & 90.1   & 92.0 & \textbf{92.6} & 92.4 & 87.8 & 90.1  \\
DTD \cite{cimpoi14describing}         & 47.6   & 49.0 & 57.8 & \textbf{58.5} & 50.7 & 53.7  \\
Aircraft \cite{maji13fine-grained}    & 31.1   & 30.1 & 38.5 & \textbf{38.5} & 29.5 & 38.0  \\
Caltech101 \cite{li_andreeto_ranzato_perona_2022}  & 71.3   & 71.9 & 75.4 & 75.5 & 70.4 & \textbf{75.6}  \\
Flowers102 \cite{Nilsback08}  & 73.0   & 76.4 & 76.6 & \textbf{76.7} & 75.5 & 74.9  \\
Food101 \cite{bossard14}     & 90.9   & 90.9 & 90.8 & 92.1 & 89.8 & \textbf{92.6}  \\
Cars \cite{KrauseStarkDengFei-Fei_3DRR2013}        &  75.9  & 76.3 & 92.9 & \textbf{93.6}  & 75.9 & 93.4 \\
Cifar10 \cite{Krizhevsky2009LearningML}     & 95.0   & 94.9 & \textbf{95.7} & 95.5  & 95.0 & 95.6 \\ 
ImageNet-1k \cite{deng2009imagenet} & 69.6   & 69.2 & 73.0 & 73.1  & 70.7 & \textbf{75.6} \\ 
\midrule 
\textbf{Vision Avg.}  & 71.6 \colorred{(-0.1)} & 72.3 \colorgreen{(+0.6)} & 77.0 \colorgreen{(+0.4)} & \textbf{77.3}  \colorgreen{(+0.7)} & 71.7 & 76.6 \\
\end{tabular}
\caption{\textbf{Zero-shot results for \modelName in object recognition tasks:} We present the performance of various LENS variations and compare them with the out-of-the-box performance of CLIP (Radford et al., 2021). In the majority of benchmarks, \modelName demonstrates competitive or superior performance compared to CLIP.}
\label{tab:vision-zero}
\end{table*}

\section{Experiments}
\label{sec:experiments}
In this section, we conduct extensive experiments and analyses to show the efficacy of \modelName. First, we compare \modelName with other state-of-the-art models \cite{radford2021learning} in object recognition. Then, we also evaluate \modelName on vision-and-language reasoning tasks and compare it to multimodal foundation models \cite{alayrac2022flamingo,huang2023language}. We also perform ablation studies on important design choices such as prompt components and prompt patterns for each task. 

\subsection{Datasets} 
\label{subsec:datasets}
For object recognition, we conduct experiments using 9 benchmark datasets presented in \cite{radford2021learning}. We examine the performance of our method in zero-shot, 1-shot, and 3-shot settings, aiming to showcase the capabilities of the frozen LLM in incorporating contextual learning \cite{brown2020language}.For vision and language reasoning, we focus on zero-shot benchmarks since we didn't see an improvement while evaluating LENS in few-shot settings. We evaluate our approach on the test-dev split of the VQA 2.0 dataset \citep{goyal2017making} and the OK-VQA dataset \citep{marino2019ok} test set. We also explore the performance of \modelName on the dev and test-seen sets of the Hateful Memes dataset \cite{kiela2020hateful} and the test set of Rendered SST2 \cite{radford2021learning}. For a detailed overview of each task, including dataset sizes, the number of classes, and the specific evaluation metrics employed, please refer to Table \ref{tab:datasets-description} in the supplementary material. 

\subsection{Implementation Details}

We use OpenCLIP-H/14\footnote{\url{https://huggingface.co/laion/CLIP-ViT-H-14-laion2B-s32B-b79K}}  and CLIP-L/14\footnote{\url{https://huggingface.co/openai/clip-vit-large-patch14}} as our default vision encoders in both tags and attributes modules. We adopt BLIP-large\footnote{\url{https://huggingface.co/Salesforce/blip-image-captioning-large}} captioning checkpoint finetuned on COCO \citep{cocodataset} in intensive captioning module. In this module, we perform a top-k sampling \cite{fan-etal-2018-hierarchical}, where k represents the desired number of captions and generates a maximum of $k=50$ captions per image. Finally, we adopt Flan-T5 models as our default family of frozen LLMs \cite{longpre2023flan}. To generate answers in line with the evaluation tasks, we employ beam search with number of beams equal to 5. Additionally, we apply a length penalty equal to -1, encouraging the generation of concise answers as in \cite{li2023blip2}. These experiments were conducted on 8 NVIDIA A100 (40GB) GPUs.

We perform task-specific optimizations on \modelName to achieve the best performance. For object recognition, we utilize the tag module, which operates on the classes corresponding to each dataset. Additionally, we employ the attribute module, which utilizes the attribute vocabulary. Based on our preliminary experiments, we skip the intensive captioning modules. In VQA tasks, we solely use the intensive captioning module, as our experiments showed that tags and captions did not provide significant improvement. For the Hateful Memes \cite{kiela2020hateful} and Rendered-SST2 datasets, we incorporate the tag, attributes, and captioning modules. We generate only one caption using beam search with a width of 5.


\subsection{Results}
\label{sec:main_results}
\begin{figure*}[t!]
  \centering
  \includegraphics[width=1\textwidth, angle=0]{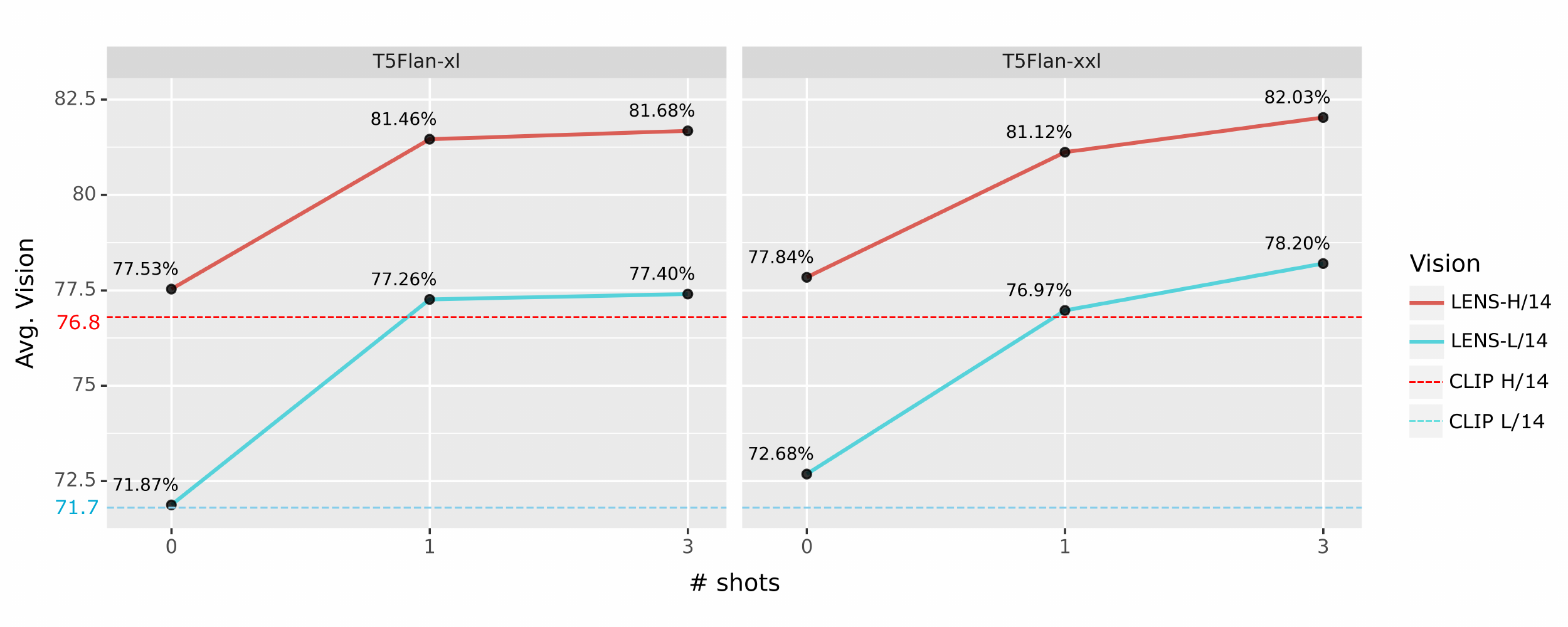}
  \caption{\textbf{Average few-shot performance of LENS on vision tasks.} We conducted tests using various flavors of CLIP vision encoders and a frozen Flan-T5 LLM. Larger LLMs provide better performance in zero-shot and three-shot settings, while larger vision encoders enhance performance across all settings.}
  \label{fig:few_shot}
\end{figure*}

We evaluate \modelName across vision and vision \& language tasks. For vision, we evaluated 8 benchmarks and compared them with state-of-the-art models in object recognition \cite{radford2021learning} in zero- and few-shot settings.  For  vision \& language, we evaluate four representative tasks for visual question answering and compare them with state-of-the-art models that employ a frozen LLM and that require additional pre-training stage(s) and big corpora of paired datasets for aligning the vision and language modalities. In these tasks, we only report zero-shot results since we do not observe an improvement while employing in-context learning.

\textbf{Object Recognition}: In Table~\ref{tab:vision-zero}, we observe that on zero-shot, \modelName composed by ViT-H/14 \cite{dosovitskiy2021image} as the visual backbone and with Flan-T5\textsubscript{xxl} as the frozen LLM outperforms in \textbf{\colorgreen{+0.7\%}} on average to equal-size CLIP which employs a common prompt. Interestingly, our experiments reveal that for object recognition tasks, there appears to be no direct relationship between the size of the frozen LLM and classification performance. However, we do observe a correspondence between the size of the tagger architecture (ViT backbone) and  performance. 

In Fig.~\ref{fig:few_shot}, we plot the average vision performance on all datasets except ImageNet (due to its large size), and observe that more shots help to increase performance under any combination of visual backbone and frozen LLM. Also, we again observe that there is no direct relationship between a better frozen LLM with respect to performance. However, we do see that a better visual backbone helps to increase the average vision performance.
\begin{table*}[!t]
\centering	
\begin{tabular}	{lc|cccccc}
\toprule	 	
 \multirow{2}{*}{Models}&\multirow{2}{*}{\makecell[l]{\# Trainable \\ Params}} &  VQAv2 & OK-VQA  & Rendered - SST2 &  \multicolumn{2}{c}{Hateful Memes}  \\
   & & test-dev & test & test & dev & test-seen\\
    \midrule
    Kosmos-1  & 1.6B    & 51.0 & -  & 67.1 & \textbf{63.9} & -\\
    Flamingo\textsubscript{3B}  & 1.4B  & 49.2 & 41.2  & - & - & 53.7\\
    Flamingo\textsubscript{9B}  & 1.8B  & 51.8 & 44.7  & - & - & 57.0 \\
    Flamingo\textsubscript{80B} & 10.2B & 56.3 & \textbf{50.6}& - & - & 46.4\\
    BLIP-2\textsubscript{ViT-L FlanT5\textsubscript{XL}} & 103M & 62.3 & 39.4 & - & - & -\\
    BLIP-2\textsubscript{ViT-g FlanT5\textsubscript{XXL}} & 108M & \textbf{65.0} & 45.9 & - & - & -\\
    \midrule
    LENS Flan-T5\textsubscript{XL} & 0 & 57.9 & 32.8  & \textbf{83.3} & 58.0 & 59.3\\
    LENS Flan-T5\textsubscript{XXL} & 0 & 62.6 & 43.3  & 82.0 & 59.4 & \textbf{62.5}\\
    \bottomrule	 
\end{tabular}
\vspace{1em}
\caption{Comparison with the state-of-the-art methods on zero-shot settings on VQAv2 \cite{goyal2017making}, OK-VQA \cite{marino2019ok}, Rendered-SST \cite{radford2021learning}, and Hateful Memes \cite{kiela2020hateful}. Trainable parameters represent the number of parameters needed for aligning the vision modality with frozen LLM. \modelName consistently outperforms or reasonably competes with extensively pretrained methods that rely on large amounts of data for multimodal alignment.}
\label{tab:vision-language-zero}
\end{table*}

\textbf{Vision and Language}: The comparative performance analysis of \modelName in relation to other systems is presented in Table~\ref{tab:vision-language-zero}. The results obtained from our experiments clearly demonstrate the highly competitive nature of \modelName, even when compared to significantly larger and more sophisticated systems such as Flamingo~\cite{alayrac2022flamingo}, BLIP-2~\cite{li2023blip2}, and Kosmos~\cite{huang2023language}.

Specifically, our findings reveal that on the VQA 2.0 \cite{goyal2017making}, \modelName Flan-T5\textsubscript{XXL} achieves superior performance over Flamingo\textsubscript{9B} and Kosmos-1 by 11\% and 15\%, respectively. Furthermore, \modelName outperforms the most powerful variant of Flamingo by 6 points. Moreover, our Intensive Captioning module, which utilizes a ViT-L vision encoder, is on par with the largest BLIP-2 architecture that employs ViT-G as its vision encoder. In addition, our best \modelName model surpasses multiple versions of Flamingo in on Hateful Memes \cite{kiela2020hateful} and exhibits better performance compared to Kosmos on the Rendered-SST2 benchmark. It is perhaps not particularly surprising that our approach does so well on Rendered SST2, where a good language model is needed to understand the sentiment of text which is extracted from an image. In this way, Rendered SST2 is not just about linking image features directly to text; it is also about interpreting what that text actually means. On OK-VQA, our model's performance does not match that of Flamingo, potentially due to the fact that the 70B Chinchilla language model using in Flamingo\textsubscript{80B} possesses a larger knowledge base than our best reasoning module, as also suggested in \cite{li2023blip2}.

\begin{wraptable}{r}{0.5\linewidth}
    \vspace{-3em}
    \centering
    \begin{tabular}{lr}
    \hline 
    Prompt Template  &  Acc. (Avg.)\\
    \hline
    Objects  &76.6\\
    Attributes & 74.7\\
    Objects + Attributes &\textbf{77.0}\\
    \hline
    \end{tabular}
    \vspace{1em}
    \caption{\textbf{Ablations on vision datasets}. We report average accuracy on the vision datasets discussed in Section~\ref{subsec:datasets}. The object information helps more than the attributes but together they are complimentary and lead to overall better performance.}
    \label{tab:vision-ablation}
    \vspace{-1em}
\end{wraptable}
\begin{figure*}[h]
  \centering
  \includegraphics[width=1\textwidth, angle=0]{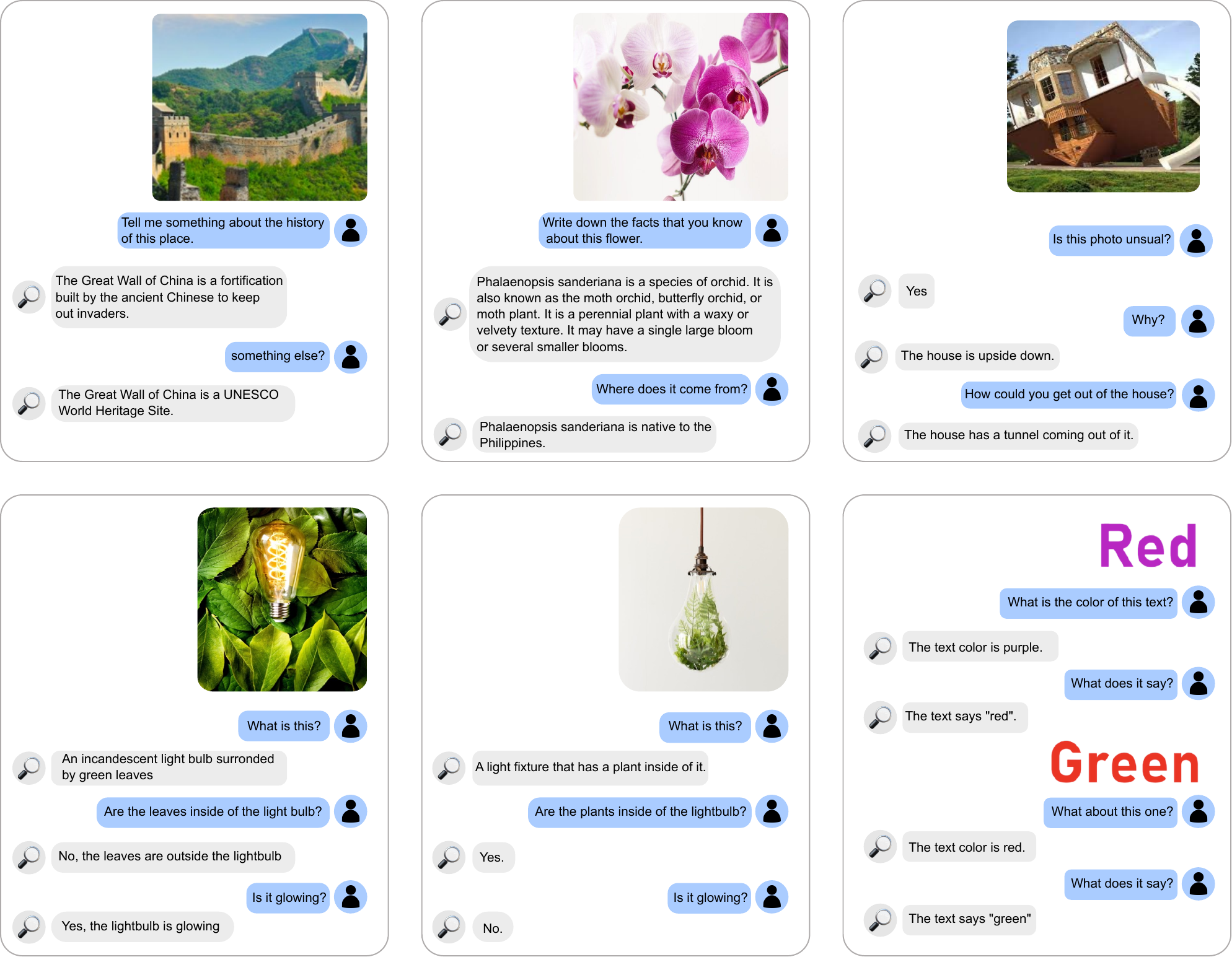}
  \caption{\textbf{Selected examples of LENS} using Tag and Attributes Modules with OpenCLIP-H/14 as the vision encoder, Intensive Captioning Module and Flan-T5\textsubscript{xxl} as the LLM.}
  \label{fig:vision-examples}
\end{figure*}

\subsection{Ablations on \modelName components }
\label{sec:ablations}
\textbf{Object recognition:} In Table~\ref{tab:vision-ablation}, we conducted an ablation study of \modelName's components on object recognition using Flan-T5\textsubscript{XL} and CLIP-H/14, as discussed in Section~\ref{sec:main_results}. We present the average accuracy across the benchmarks. By utilizing only the tag module, we exclusively rely on CLIP and observe similar performance as CLIP-H/14 in Table~\ref{tab:vision-zero}. However, we noticed a drop in performance when using only the attributes module. When combined with tags, attributes significantly contribute to enhancing the performance of \modelName by \textbf{+0.4\%}. This demonstrates that LENS serves as a robust baseline compared to using only a single vision module such as CLIP. For a detailed overview on each dataset, please refer to Table~\ref{tab:ablations-vision-detailed} in supplementary material.

\textbf{Visual Reasoning}: For the VQA 2.0 dataset (Goyal et al., 2017), we conducted ablations using our model name, which is equipped with Flan-T5\textsubscript{XXL}, on the minival split. As shown in Table~\ref{tab:vision-language-ablation}, we noticed that increasing the number of captions generated by our Intensive Captioning module led to a gradual improvement in performance. However, it eventually reaches a saturation point, indicating that the module provides valuable information about the image only up to a certain threshold.

We also conducted ablation studies on the \modelName components using the dev set of the Hateful-Memes benchmark \cite{kiela2020hateful}. Table~\ref{tab:ablation-hm} demonstrates that a combination of the global captioning, tags, and attributes modules is essential for achieving high performance on this task. Specifically, we observed that both tags and attributes contribute more to the performance improvement compared to the global captioning module when combined with OCR. However, it is important to note that all of these components are necessary and their combined usage results in the best performance. We also present several qualitative examples from \modelName in Fig.~\ref{fig:vision-examples}, illustrating its reasoning capabilities by answering questions about complex scenes and scenarios.

\begin{table}
\centering
\parbox{.45\linewidth}{
    \centering
    \begin{tabular}{lr}
    \hline 
    Prompt Template & ROC-AUC\\
    \hline
    OCR& 57.2\\
    Objects + OCR& 58.4\\
    Attributes + OCR& 59.3\\
    Caption + OCR& 57.2\\
    All & 59.4\\
    \hline
    \end{tabular}
    \vspace{1em}
    \caption{\textbf{Hateful Memes \cite{kiela2020hateful} ablations.} Adding more visual information on top of OCR improves the performance consistently though attributes help the most.}
    \label{tab:ablation-hm}
}
\hfill
\parbox{.50\linewidth} {
    \centering
    \begin{tabular}{lr}
    \hline 
    Prompt Template & VQA-ACC\\
    \hline
    Question & 37.2\\
    Intensive Captioning (1)  + Question  & 52.5\\
    Intensive Captioning (5)  + Question & 56.6\\
    Intensive Captioning (20)  + Question  & 59.1\\
    Intensive Captioning (50)  + Question & \textbf{60.4}\\
    \hline
    \end{tabular}
    \vspace{1em}
    \caption{Ablation results on VQA 2.0 \cite{goyal2017making}. Increasing the number of intensive captions improves the performance of \modelName gradually on VQA but starts saturating eventually.}
    \label{tab:vision-language-ablation}
}
\end{table}



\section{Conclusion}
\label{sec:conclusion}
We introduce \modelNameWithEmoji, a generic and computationally efficient method that enables a frozen LLM to effectively coordinate vision modules, resulting in competitive performance even when compared to larger multimodally pretrained systems. \modelName offers adaptability to various open-source or black-box language models, regardless of their pretraining or multimodal data, thereby providing flexibility and scalability for future improvements in performance within the community.

By leveraging the strengths of LLMs and our modular approach, \modelName represents a significant advancement in task-solving without the need for additional pretraining. Its seamless integration with diverse vision tasks showcases its versatility and potential for widespread application.

In future work, an intriguing direction to explore would be expanding the applicability of \modelName by incorporating it into tasks involving different modalities. For instance, integrating \modelName into audio classification or video action reasoning tasks could yield valuable insights. This expansion would involve orchestrating the roles of the LLM and integrating it with complementary modules.
\section{Limitations}
\label{sec:limitation}
As with any research work, \modelName has its own limitations. We aim to address a few of them in this section. Firstly, the vision capability of \modelName heavily relies on its underlying vision components, namely CLIP and BLIP. Although these models have shown notable performance improvements, there is still room for further enhancement by leveraging their strengths and combining them with LLMs. We demonstrate a few failure cases of LENS in Fig.~\ref{fig:incorrec-examples} in the supplementary material.  Future research should explore methods to effectively integrate these models and harness the synergies between vision and language components to achieve even better performance across diverse tasks.

Secondly, it is important to acknowledge that conducting evaluation experiments with \modelName models requires substantial computational resources. For example, our experiments were conducted using 8*A100, which may pose challenges for smaller or medium-sized labs, as well as underrepresented communities with limited access to such resources. However, it is worth noting that the computational costs associated with evaluating \modelName models are comparatively lower than the extensive training requirements of larger visual-language models, such as Flamingo, which can demand upwards of 500k TPU hours. Nonetheless, efforts should be made to make computational resources more accessible and explore methods for reducing the computational burden while maintaining the effectiveness of \modelName.

\section*{Acknowledgments}
\label{sec:acknowledge}
We would like to express our gratitude to the Fatima Fellowship and Hugging Face for granting computational resources for the preliminary experiments of this project.
\clearpage
\bibliography{neurips_2023}
\bibliographystyle{neurips_2023}
\clearpage

\appendix
\begin{center}
\textbf{\Large Supplementary Material }
\end{center}
\section{\modelName datasets}

In Table~\ref{tab:datasets-description}, we present the statistics for each of the datasets utilized in the evaluation of \modelName. We provide details regarding the evaluation metric employed, as well as whether the evaluation was conducted in an open-ended manner or a closed-ended manner with a fixed class vocabulary.
\begin{table*}[h]
\centering
\footnotesize
\begin{tabular}{lcccc}

\toprule
     Dataset & Split(s) & Size & Evaluation Method & Metric\\
    \toprule
     \emph{Image Classification} & &\\
    Oxford-IIIT Pets \cite{parkhi2012pets} & test &  3,669 & close-ended & mean per class\\
    Describable Textures \cite{cimpoi14describing} & test & 1,880  & close-ended & accuracy\\
    Caltech-101 \cite{FeiFei2004LearningGV} & test &  6,085 &  close-ended & accuracy\\
    Oxford Flowers 102 \cite{Nilsback08} & test & 6,149 & close-ended & mean per class\\
    FGVC Aircraft \cite{maji13fine-grained} & test & 3,333 & close-ended & mean per class\\
    Food101 \cite{bossard14} & test & 25,250 & close-ended & accuracy\\
    Cifar10 \cite{bossard14} & test & 10,000 & close-ended & accuracy\\
    ImageNet-1k \cite{bossard14} & validation & 50,000 & close-ended & accuracy\\
    \midrule
    \midrule
    \emph{Vision \& Language} & &\\
    Hateful Memes \cite{kiela2020hateful} & dev & 500 & open-ended & ROC AUC\\
    Hateful Memes \cite{kiela2020hateful} & test-seen & 1,000 & open-ended & ROC AUC\\
    VQA 2.0 \cite{goyal2017making}  & tesdev &  107,394 & open-ended & VQA accuracy\\
    OK-VQA \cite{marino2019ok} & validation & 5,046 & open-ended & VQA accuracy\\
    Rendered SST2 \cite{marino2019ok} & validation & 1,821 & open-ended & VQA accuracy\\
    \toprule
    \end{tabular}
    \caption{\textbf{Datasets examined for evaluation of \modelName.} We describe the size statistics, the split, the method of evaluation, and metrics used for each of these datasets.}
    \label{tab:datasets-description}
\end{table*}.

\section{Detailed Ablations in Object Recognition}
In Table~\ref{tab:ablations-vision-detailed}, we present a detailed ablation result in the object recognition task presented in Section~\ref{sec:main_results}
\begin{table*}[h]
\centering
\footnotesize
\begin{tabular}{lcccc}

\toprule
Dataset    & Objects & Attributes & Objects + Attributes\\
\midrule 
Pets \cite{parkhi2012pets}     & 90.1   & 91.0 & \textbf{92.6}  \\
DTD \cite{cimpoi14describing}    & 53.7   & 51.5 & \textbf{57.8} \\
Aircraft \cite{maji13fine-grained} & 38.0  & 36.5 & \textbf{38.5}  \\
Caltech101 \cite{li_andreeto_ranzato_perona_2022} & \textbf{75.6} & 71.6 & 75.4 \\
Flowers102 \cite{Nilsback08}& 74.9 & 75.6 & \textbf{76.6} \\
Food101 \cite{bossard14}  & \textbf{92.6} & 89.1 & 90.8 \\
Cars \cite{KrauseStarkDengFei-Fei_3DRR2013}  & \textbf{93.4}    & 92.1 & 92.9 \\
Cifar10 \cite{Krizhevsky2009LearningML}  & 95.6  & 93.4& \textbf{95.7} \\
ImageNet-1k \cite{deng2009imagenet} & \textbf{75.6}  & 71.5 & 73.0 \\
\midrule 
\textbf{Avg. Vision} & 76.8 & 75.1 & \textbf{77.5}  \\
\toprule

\end{tabular}
\caption{Ablations results using OpenCLIP-H/14 as vision encoder and Flan-T5\textsubscript{XL} as the LLM }
\label{tab:ablations-vision-detailed}
\end{table*}

\begin{table}[htbp]
  \centering
  \begin{minipage}[t]{0.45\textwidth}
    \centering
  \begin{tabular}{p{2.5cm}p{3.5cm}}
  \toprule
    Components & Prompt \\
    \midrule
    Tag: & \texttt{Top-1 CLIP Tag}\\
    \midrule
    Attributes: &  \texttt{Top-K Attributes}\\
    \toprule
    Question: &  \texttt{Task specific prompt}\\
    \toprule
    Short Answer: & \texttt{\{answer\}}\\
    \toprule
  \end{tabular}
  \caption{Object recognition prompt used in LENS}
  \end{minipage}\hfill
  \begin{minipage}[t]{0.45\textwidth}
    \centering
  \begin{tabular}{p{2.5cm}p{3.5cm}}
  \toprule
    Components & Prompt \\
    \midrule
    Captions: & \texttt{Top-N captions}\\
    \midrule
    Question: &  \texttt{e.g Who is doing "x" action?}\\
    \toprule
    Short Answer: & \texttt{\{answer\}}\\
    \toprule
  \end{tabular}
  \caption{VQA prompt used in LENS}
  \end{minipage}
  
  \vspace{1cm} 
  
  \begin{minipage}[t]{0.45\textwidth}
    \begin{tabular}{p{2.2cm}p{3.2cm}}
  \toprule
    Components & Prompt \\
    \midrule
    Attributes: & \texttt{Top-K Attributes tags}\\
    \midrule
    Tags: & \texttt{Top-K Object tags}\\
    \midrule
    Caption: & \texttt{BLIP-Caption}\\
    \midrule
    OCR: & \texttt{OCR Meme text}\\
    \midrule
    Question: &  \texttt{Is the image hateful or not hateful?}\\
    \toprule
    Short Answer: & \texttt{\{answer\}}\\
    \toprule
  \end{tabular}
  \caption{Hateful-memes prompt used in LENS}
  \end{minipage}
\end{table}

\section{Failure examples}
\begin{figure*}[h!]
  \centering
  \includegraphics[width=0.52\textwidth, angle=0]{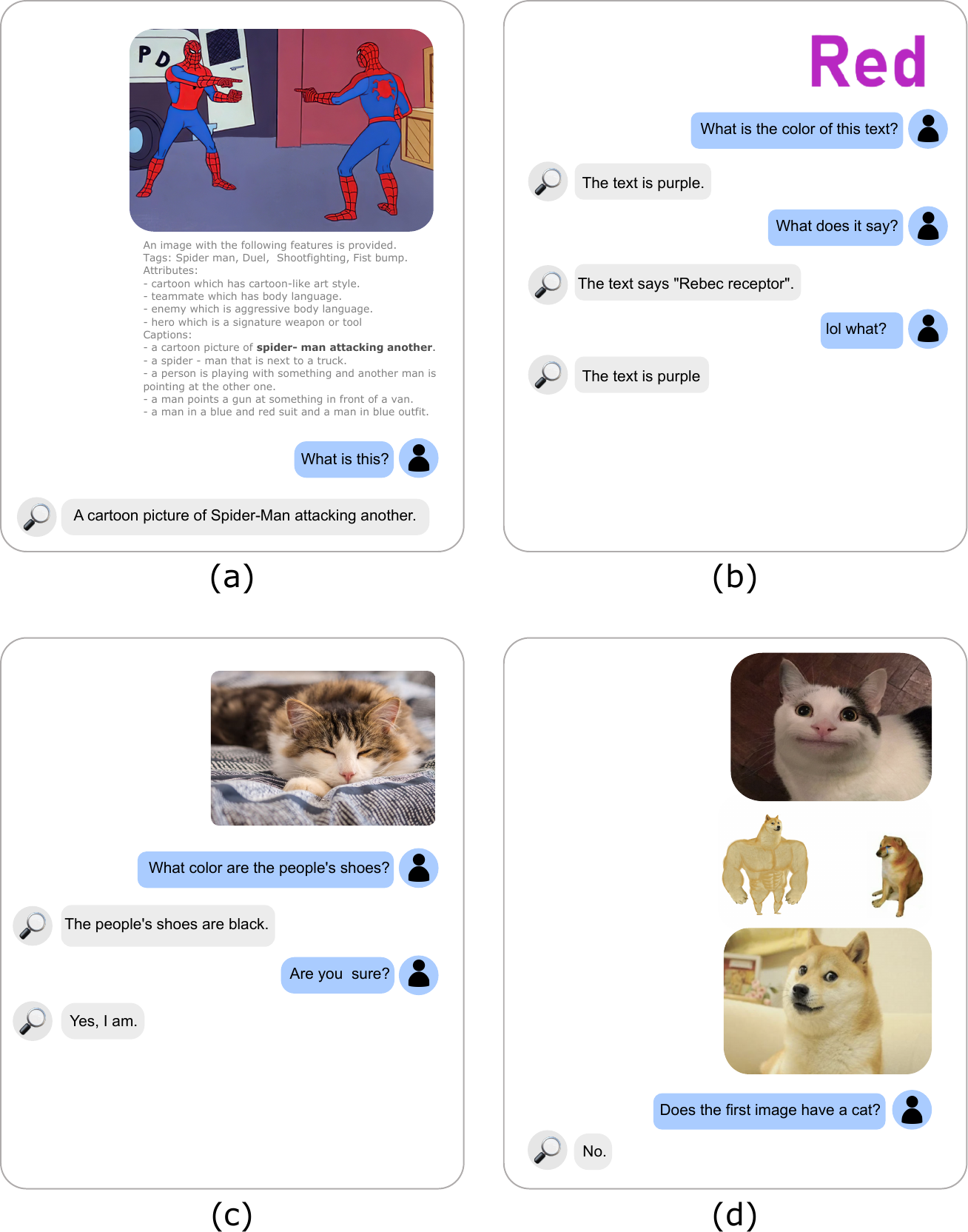}
  \caption{\textbf{Incorrect outputs of \modelName} using Tag and Attributes Modules with OpenCLIP-H/14 as the vision encoder, Intensive Captioning Module and Flan-T5\textsubscript{xxl} as the LLM. (a) Incorrect Visual Information (b) Inconsistency (c) Presuppositions (d) Forgetting.}
  \label{fig:incorrec-examples}
\end{figure*}

In Fig.~\ref{fig:incorrec-examples}, we showcase some of the failure cases for our \modelName. These involve (i) failures from vision modules to correctly identify objects or attributes (ii) inconsistency between the responses (iii) presuppositions and in-built biases and (iv) forgetting and limitations of context windows of the LLMs.

\end{document}